\documentclass[10pt,twocolumn,letterpaper]{article}

\usepackage{cvpr}
\usepackage{times}
\usepackage{epsfig}
\usepackage{graphicx}
\usepackage{amsmath}
\usepackage{amssymb}

\usepackage{color}
\usepackage{xcolor}
\usepackage[linesnumbered,ruled,vlined]{algorithm2e}
\usepackage{booktabs} %
\usepackage{caption}
\usepackage{subfig}

\usepackage{float}
\usepackage{aliascnt}
\usepackage{dsfont}
\usepackage{multirow}
\usepackage{eqparbox}
\usepackage{tablefootnote}
\usepackage{enumitem}

\usepackage{pdfpages}

\usepackage[pagebackref=true,breaklinks=true,letterpaper=true,colorlinks,bookmarks=false]{hyperref}

\cvprfinalcopy %

\SetCommentSty{mycommfont}

\SetKwInput{KwInput}{Input}                %
\SetKwInput{KwOutput}{Output}              %

\newtheorem{theorem}{Theorem}

\newcommand{\argmin}{\mathop{\mathrm{argmin}}} 
\newcommand{\ignore}[1]{}

\DeclareRobustCommand\code[1]{%
  \ifmmode
    \expandafter\texttt
  \else
    \expandafter\textnhtt
  \fi{#1}%
}

\newaliascnt{eqfloat}{equation}
\newfloat{eqfloat}{h}{eqflts}
\floatname{eqfloat}{Equation}

\newcommand*{\ORGeqfloat}{}
\let\ORGeqfloat\eqfloat
\def\eqfloat{%
  \let\ORIGINALcaption\caption
  \def\caption{%
    \addtocounter{equation}{-1}%
    \ORIGINALcaption
  }%
  \ORGeqfloat
}

\def\cvprPaperID{826} %

\ifcvprfinal\fi
\begin{document}

\title{Deep Global Registration}

\author{Christopher Choy\thanks{indicates equal contribution}\\
Stanford University\\
\and
Wei Dong${}^*$\\
Carnegie Mellon University\\
\and
Vladlen Koltun\\
Intel Labs\\
}

\maketitle

\begin{abstract}
We present Deep Global Registration, a differentiable framework for pairwise registration of real-world 3D scans. Deep global registration is based on three modules: a 6-dimensional convolutional network for correspondence confidence prediction, a differentiable Weighted Procrustes algorithm for closed-form pose estimation, and a robust gradient-based $\text{SE}(3)$ optimizer for pose refinement. Experiments demonstrate that our approach outperforms state-of-the-art methods, both learning-based and classical, on real-world data.
\end{abstract}

\section{Introduction}
\label{sec:intro}

A variety of applications, including 3D reconstruction, tracking, pose estimation, and object detection, invoke 3D registration as part of their operation~\cite{rusu2009icra, cai20103d, qian2014realtime}. To maximize the accuracy and speed of 3D registration, researchers have developed geometric feature descriptors~\cite{cgf, ppf_fold, wang2019deep, FCGF2019}, pose optimization algorithms~\cite{schnabel2007efficient, yang2015go, lucas1981iterative, zhou2016eccv}, and end-to-end feature learning and registration pipelines~\cite{wang2019deep, aoki2019pointnetlk}.

\begin{figure}[t]
    \small 
    \centering
    \subfloat[RANSAC~\cite{rusu2009icra}]{
    \includegraphics[width=.9\linewidth]{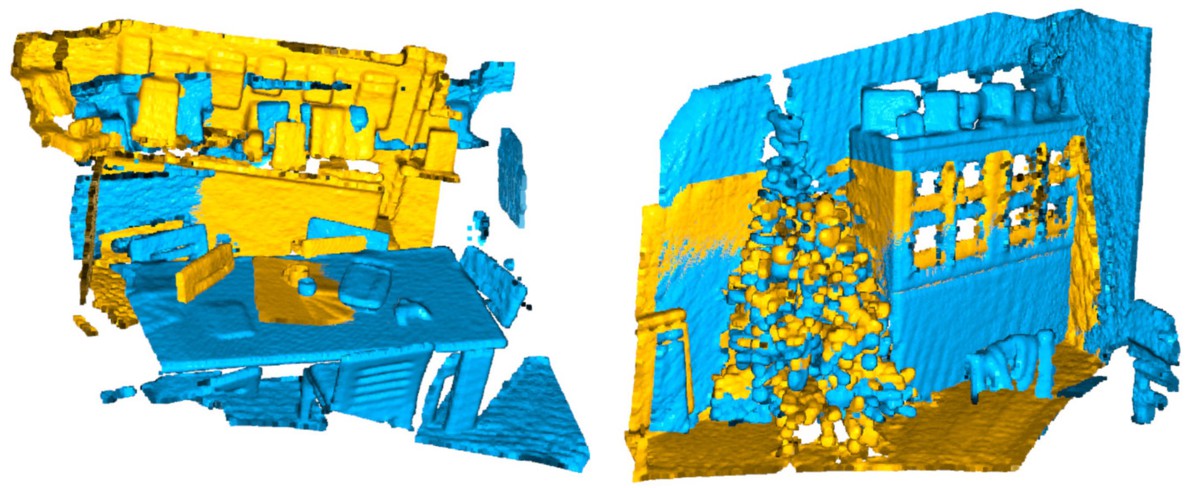}
    }\\
    \vspace{-1.5em}
    \subfloat[FGR~\cite{zhou2016eccv}]{
    \includegraphics[width=.9\linewidth]{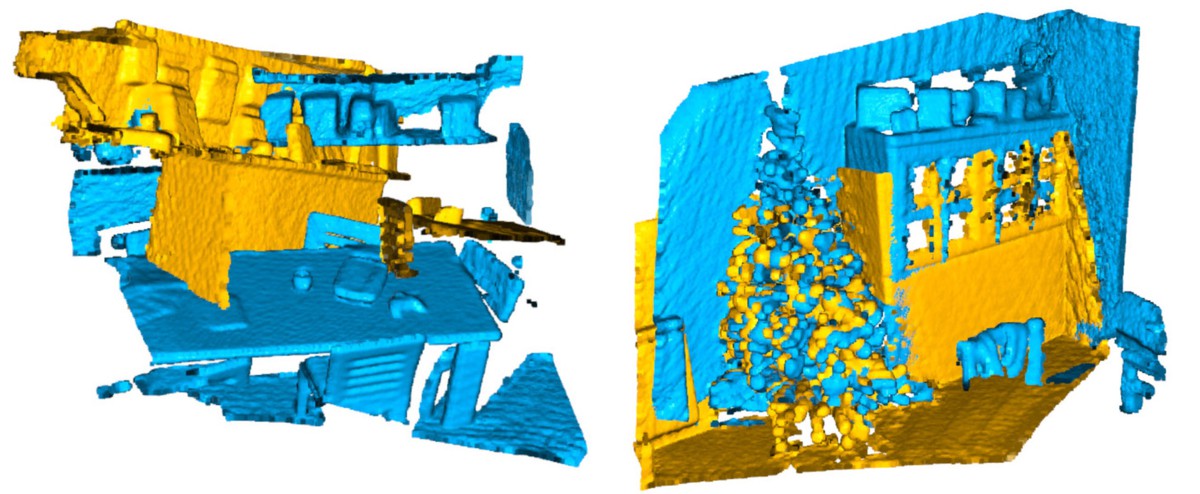}
    }\\
    \vspace{-1.5em}
    \subfloat[DCP~\cite{wang2019deep}]{
    \includegraphics[width=.9\linewidth]{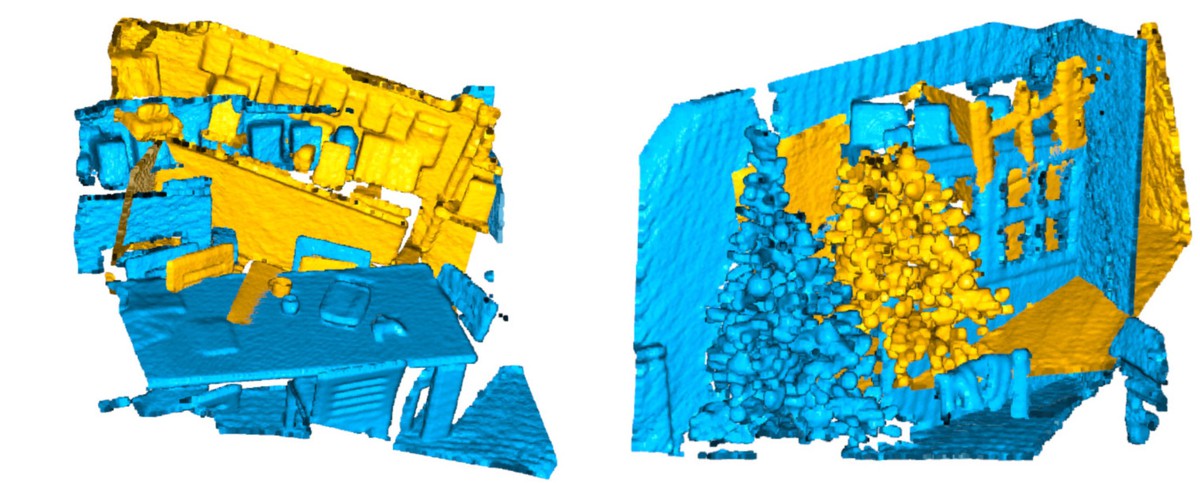}
    }\\
    \vspace{-1.5em}
    \subfloat[Ours]{
    \includegraphics[width=.9\linewidth]{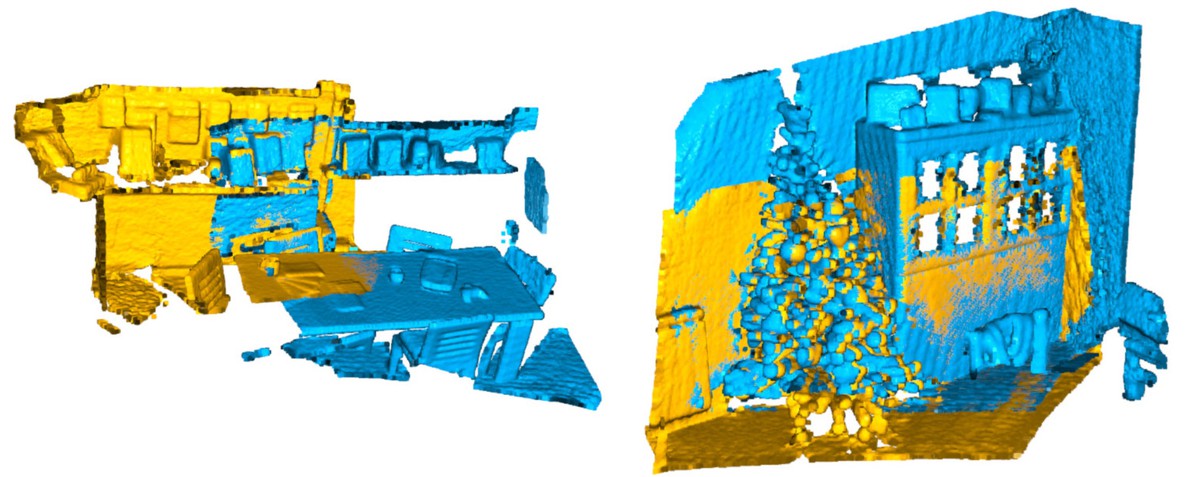}
    }
  \vspace{-0.5em}
  \caption{Pairwise registration results on the 3DMatch dataset~\cite{zeng20163dmatch}. Our method successfully aligns a challenging 3D pair (\textit{left}), while RANSAC~\cite{rusu2009icra}, FGR~\cite{zhou2016eccv}, and DCP~\cite{wang2019deep} fail. On an easier pair (\textit{right}), our method achieves finer alignment.}
  \label{fig:teaser}
  \vspace{-1.5em}
\end{figure}

In particular, recent end-to-end registration networks have proven to be effective in relation to classical pipelines.
However, these end-to-end approaches have some drawbacks that limit their accuracy and applicability.
For example, PointNetLK~\cite{aoki2019pointnetlk} uses globally pooled features to encode the entire geometry of a point cloud, which decreases spatial acuity and registration accuracy.
Deep closest point~\cite{wang2019deep} makes strong assumptions on the distribution of points and correspondences, which do not hold for partially overlapping 3D scans.

In this work, we propose three modules for robust and accurate registration that resolve these drawbacks: a 6-dimensional convolutional network for correspondence confidence estimation, a differentiable Weighted Procrustes method for scalable registration, and a robust $\text{SE}(3)$ optimizer for fine-tuning the final alignment.

The first component is a 6-dimensional convolutional network that analyzes the geometry of 3D correspondences and estimates their accuracy.
Our approach is inspired by a number of learning-based methods for estimating the validity of correspondences in 2D~\cite{Yi2018,Ranftl2018} and 3D~\cite{dias2019corr}. 
These methods stack coordinates of correspondence pairs, forming a vector $[\mathbf{x}; \mathbf{y}] \in \mathbb{R}^{2 \times D}$ for each correspondence $\mathbf{x}, \mathbf{y} \in \mathbb{R}^D$.
Prior methods treat these $2\times D$-dimensional vectors as a set, and apply global set processing models for analysis. Such models largely disregard local geometric structure. Yet the correspondences are embedded in a metric space ($\mathbb{R}^{2 \times D}$) that induces distances and neighborhood relationships.
In particular, 3D correspondences form a geometric structure in 6-dimensional space~\cite{choy2020high} and we use a high-dimensional convolutional network to analyze the 6D structure formed by correspondences and estimate the likelihood that a given correspondence is correct (\ie, an inlier).

The second component we develop is a differentiable Weighted Procrustes solver. The Procrustes method~\cite{gower1975generalized} provides a closed-form solution for rigid registration in $\text{SE}(3)$. A differentiable version of the Procrustes method by Wang \etal~\cite{wang2019deep} has been used for end-to-end registration.
However, the differentiable Procrustes method passes gradients through coordinates, which requires $O(N^2)$ time and memory for $N$ keypoints, limiting the number of keypoints that can be processed by the network.
We use the inlier probabilities predicted by our first module (the 6D convolutional network) to guide the Procrustes method, thus forming a differentiable Weighted Procrustes method. This method passes gradients through the weights associated with correspondences rather than correspondence coordinates.
The computational complexity of the Weighted Procrustes method is linear in the number of correspondences, allowing the registration pipeline to use dense correspondence sets rather than sparse keypoints. This substantially increases registration accuracy.

Our third component is a robust optimization module that fine-tunes the alignment produced by the Weighted Procrustes solver.
This optimization module minimizes a differentiable loss via gradient descent on the continuous $\text{SE}(3)$ representation space~\cite{zhou2019continuity}.
The optimization is fast since it does not require neighbor search in the inner loop~\cite{zhang1994iterative}.

Experimentally,
we validate the presented modules on a real-world pairwise registration benchmark~\cite{zeng20163dmatch} and large-scale scene reconstruction datasets~\cite{ahanda2014icra,choi2015cvpr,Park2017}.
We show that our modules are robust, accurate, and fast in comparison to both classical global registration algorithms~\cite{zhou2016eccv,rusu2009icra,yang2015go} and recent end-to-end approaches~\cite{dias2019corr, wang2019deep, aoki2019pointnetlk}.
All training and experiment scripts are available at~{\small \url{https://github.com/chrischoy/DeepGlobalRegistration}}.

\section{Related Work}

We divide the related work into three categories following the stages of standard registration pipelines that deal with real-world 3D scans: feature-based correspondence matching, outlier filtering, and pose optimization.

\noindent \textbf{Feature-based correspondence matching.}
The first step in many 3D registration pipelines is feature extraction. Local and global geometric structure in 3D is analyzed to produce high-dimensional feature descriptors, which can then be used to establish correspondences.

Traditional hand-crafted features commonly summarize pairwise or higher-order relationships in histograms~\cite{johnson1999spin, shot, usc, pfh, rusu2009icra}.
Recent work has shifted to learning features via deep networks~\cite{zeng20163dmatch,cgf}. A number of recent methods are based on global pooling models~\cite{ppf, ppf_fold, Zhao2019}, while others use convolutional networks~\cite{perfectmatch, FCGF2019}.

Our work is agnostic to the feature extraction mechanism. Our modules primarily address subsequent stages of the registration pipeline and are compatible with a wide variety of feature descriptors.

\noindent \textbf{Outlier filtering.}
Correspondences produced by matching features are commonly heavily contaminated by outliers.
These outliers need to be filtered out for robust alignment.
A widely used family of techniques for robust model fitting is based on RANdom SAmple Consensus (RANSAC)~\cite{schnabel2007efficient,aiger20084pts, rusu2009icra, mellado2014super, holz2015registration}, which iteratively samples small sets of correspondences in the hope of sampling a subset that is free from outliers.
Other algorithms are based on branch-and-bound~\cite{yang2015go}, semi-definite programming~\cite{maron2016point,maciel2003global}, and maximal clique selection~\cite{yang2019polynomial}.
These methods are accurate, but commonly require longer iterative sampling or more expensive computation as the signal-to-noise ratio decreases. One exception is TEASER~\cite{yang2019polynomial}, which remains effective even with high outlier rates.
Other methods use robust loss functions to reject outliers during optimization~\cite{zhou2016eccv, bouaziz2013sparse}.

Our work uses a convolutional network to identify inliers and outliers. The network needs only one feed-forward pass at test time and does not require iterative optimization.

\noindent \textbf{Pose optimization.}
Pose optimization is the final stage that minimizes an alignment objective on filtered correspondences.
Iterative Closest Points (ICP)~\cite{besl1992pami} and Fast Global Registration (FGR)~\cite{zhou2016eccv} use second-order optimization to optimize poses. Makadia~\etal~\cite{makadia2006fully} propose an iterative procedure to minimize correlation scores.
Maken~\etal~\cite{maken2019speeding} propose to accelerate this process by stochastic gradient descent.

Recent end-to-end frameworks combine feature learning and pose optimization. Aoki \etal.~\cite{aoki2019pointnetlk} combine PointNet global features with an iterative pose optimization method~\cite{lucas1981iterative}. Wang \etal.~\cite{wang2019deep, wang2019nips} train graph neural network features by backpropagating through pose optimization. %

We further advance this line of work. In particular, our Weighted Procrustes method reduces the complexity of optimization from quadratic to linear and enables the use of dense correspondences for highly accurate registration of real-world scans.

\section{Deep Global Registration}

3D reconstruction systems typically take a sequence of partial 3D scans as input and recover a complete 3D model of the scene. These partial scans are scene fragments, as shown in Fig.~\ref{fig:teaser}. In order to reconstruct the scene, reconstruction systems often begin by aligning pairs of fragments~\cite{choi2015cvpr}. This stage is known as pairwise registration. The accuracy and robustness of pairwise registration are critical and often determine the accuracy of the final reconstruction.

Our pairwise registration pipeline begins by extracting pointwise features. These are matched to form a set of putative correspondences. We then use a high-dimensional convolutional network (ConvNet) to estimate the veracity of each correspondence. Lastly, we use a Weighted Procrustes method to align 3D scans given correspondences with associated likelihood weights, and refine the result by optimizing a robust objective.

The following notation will be used throughout the paper. We consider two point clouds, $X = [\mathbf{x}_1, ..., \mathbf{x}_{N_x}] \in \mathbb{R}^{3 \times N_x}$ and $Y = [\mathbf{y}_1, ..., \mathbf{y}_{N_y}] \in \mathbb{R}^{3 \times N_y}$, with $N_x$ and $N_y$ points respectively, where $\mathbf{x}_i, \mathbf{y}_j \in \mathbb{R}^3$. A correspondence between $\mathbf{x}_i$ and $\mathbf{y}_j$ is denoted as $\mathbf{x}_i \leftrightarrow \mathbf{y}_j$ or $(i, j)$.

\subsection{Feature Extraction}
\label{sec:features}

To prepare for registration, we extract pointwise features that summarize geometric context in the form of vectors in metric feature space. Our pipeline is compatible with many feature descriptors. We use Fully Convolutional Geometric Features (FCGF)~\cite{FCGF2019}, which have recently been shown to be both discriminative and fast. FCGF are also compact, with dimensionality as low as 16 to 32, which supports rapid neighbor search in feature space.

\subsection{Correspondence Confidence Prediction}
\label{sec:6dconvnet}

Given the features $\mathcal{F}_x = \{ \mathbf{f}_{\mathbf{x}_1}, ..., \mathbf{f}_{\mathbf{x}_{N_x}}\}$ and $\mathcal{F}_y = \{ \mathbf{f}_{\mathbf{y}_1}, ..., \mathbf{f}_{\mathbf{y}_{N_x}}\}$ of two 3D scans, we use the nearest neighbor in the feature space to generate a set of putative correspondences or matches $\mathcal{M} = \{(i, \argmin_j \|\mathbf{f}_{\mathbf{x}_i} - \mathbf{f}_{\mathbf{y}_j}\|) | i \in [1, ..., N_x]\}$.
This procedure is deterministic and can be hand-crafted to filter out noisy correspondences with ratio or reciprocity tests~\cite{zhou2016eccv}.
However, we propose to learn this heuristic filtering process through a convolutional network that learns to analyze the underlying geometric structure of the correspondence set.

We first provide a 1-dimensional analogy to explain the geometry of correspondences.
Let $A$ be a set of 1-dimensional points $A = \{0, 1, 2, 3, 4\}$ and $B$ be another such set $B = \{10, 11, 12, 13, 14\}$. Here $B$ is a translation of $A$: $B = \{a_i + 10 | a_i \in A\}$. If an algorithm returns a set of possible correspondences $\{(0, 10), (1, 11), (2, 12), (3, 13), (4, 14), (0, 14), (4, 10)\}$, then the set of correct correspondences (inliers) will form a line (first 5 pairs), whereas incorrect correspondences (outliers) will form random noise outside the line (last 2 pairs).
If we extend this to 3D scans and pointclouds, we can also represent a 3D correspondence $\mathbf{x}_i \leftrightarrow \mathbf{y}_j$ as a point in 6-dimensional space $[\mathbf{x}_i^T, \mathbf{y}_j^T]^T \in \mathbb{R}^6$. The inlier correspondences will be distributed on a lower-dimensional surface in this 6D space, determined by the geometry of the 3D input.
We denote $\mathcal{P} = \{(i,j)| \; \|T^*(\mathbf{x}_i) - \mathbf{y}_j\| < \tau, (i,j) \in \mathcal{M} \}$ as a set of inliers or a set of correspondences $(i,j)$ that align accurately up to the threshold $\tau$ under the ground truth transformation $T^*$.
Meanwhile, the outliers $\mathcal{N} = \mathcal{P}^C \cap \mathcal{M}$ will be scattered outside the surface $\mathcal{P}$. To identify the inliers,
we use a convolutional network. Such networks have been proven effective in related dense prediction tasks, such as 3D point cloud segmentation~\cite{Graham2018,choy20194d}.
The convolutional network in our setting is in 6-dimensional space~\cite{choy2020high}. The network predicts a likelihood for each correspondence, which is a point in 6D space $[\mathbf{x}_i^T, \mathbf{y}_j^T]^T$. The prediction is interpreted as the likelihood that the correspondence is true: an inlier.

\begin{figure}
\centering
\small
\includegraphics[width=.99\linewidth]{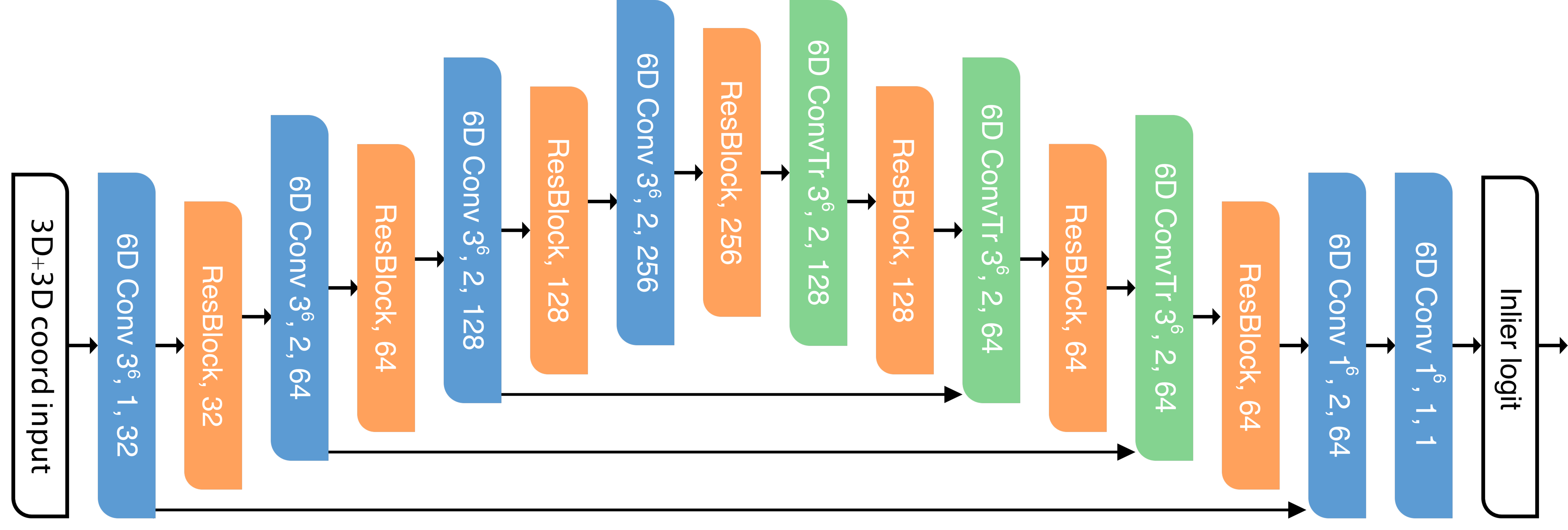}
\caption{6-dimensional convolutional network architecture for inlier likelihood prediction (Sec.~\ref{sec:6dconvnet}). The network has a U-net structure with residual blocks between strided convolutions. Best viewed on the screen.}
\label{fig:network}  
\end{figure}

Note that the convolution operator is translation invariant, thus our 6D ConvNet will generate the same output regardless of the absolute position of inputs in 3D. We use a similar network architecture to Choy et al.~\cite{FCGF2019} to create a 6D convolutional network with skip connections within the spatial resolution across the network.
The architecture of the 6D ConvNet is shown in Fig.~\ref{fig:network}.
During training, we use the binary cross-entropy loss between the likelihood prediction that a correspondence $(i,j)$ is an inlier, $p_{(i,j)} \in [0,1]$, and the ground-truth correspondences $\mathcal{P}$ to optimize the network parameters:
\begin{equation}
\small
L_\text{bce}(\mathcal{M}, T^*) = \frac{1}{|\mathcal{M}|}\bigg(\sum_{(i,j) \in \mathcal{P}} \log p_{(i,j)} + \sum_{(i,j) \in \mathcal{N}} \log p_{(i,j)}^C\bigg),
\end{equation}
where $p^C = 1 - p$ and $|\mathcal{M}|$ is the cardinality of the set of putative correspondences.

\subsection{Weighted Procrustes for $\text{SE}(3)$}
\label{sec:weighted_procrustes}

The inlier likelihood estimated by the 6D ConvNet provides a weight for each correspondence.
The original Procrustes method~\cite{gower1975generalized} minimizes the mean squared error between corresponding points $\frac{1}{N}\sum_{(i,j) \in \mathcal{M}} \| \mathbf{x}_i - \mathbf{y}_j\|^2$ and thus gives equal weight to all correspondences. In contrast, we minimize a weighted mean squared error $\sum_{(i,j) \in \mathcal{M}} w_{(i,j)} \| \mathbf{x}_i - \mathbf{y}_j\|^2$.
This change allows us to pass gradients through the weights, rather than through the position~\cite{wang2019deep}, and enables the optimization to scale to dense correspondence sets.

Formally, Weighted Procrustes analysis minimizes:
\begin{align}
    e^2 & = e^2(R, \mathbf{t}; \mathbf{w}, X, Y) \\
        & = \sum_{(i,j)\in \mathcal{M}} \tilde{w}_{(i,j)}(\mathbf{y}_j - (R\mathbf{x}_i + \mathbf{t}))^2 \label{eq:mse_w}\\
        & = \text{Tr}\left((Y - RX - \textbf{t}\textbf{1}^T)W(Y - RX - \textbf{t}\textbf{1}^T)^T\right), \label{eq:mse}
\end{align}
where $\mathbf{1} = (1, ..., 1)^T$, $X = [\mathbf{x}_1, ..., \mathbf{x}_{|\mathcal{M}|}]$, and $Y = [\mathbf{y}_{J_1}, ..., \mathbf{y}_{J_{|\mathcal{M}|}}]$.
$J$ is a list of indices that defines the correspondences $\mathbf{x}_i \leftrightarrow \mathbf{y}_{J_i}$. 
$\mathbf{w} = [w_1, \cdots, w_{|\mathcal{M}|}]$ is the weight vector and $\tilde{\mathbf{w}} =  [\tilde{w}_1, \cdots, \tilde{w}_{|\mathcal{M}|}] \triangleq \frac{\phi(\mathbf{w})}{||\phi(\mathbf{w})||_1}$ denotes the normalized weight after a nonlinear transformation $\phi$ that applies heuristic prefiltering. $W = \text{diag}(\tilde{\mathbf{w}})$ forms the diagonal weight matrix.

\begin{theorem}: The $R$ and $\textbf{t}$ that minimize the squared error $e^2(R, \textbf{t}) =\sum_{(i,j)} w_{(i,j)} (\textbf{y}_j - R\textbf{x}_i - \textbf{t})^2$ are $\hat{\textbf{t}} = (Y - RX) W\mathbf{1}$ and $\hat{R} = USV^T$ where $U\Sigma V^T = \text{SVD}(\Sigma_{xy})$, $\Sigma_{xy} = YKWKX^T$, $K = I - \sqrt{\tilde{\mathbf{w}}}\sqrt{\tilde{\mathbf{w}}}^T$, and $S = \text{diag}\left(1, \cdots, 1, \text{det}(U)\text{det}(V)\right)$.
\end{theorem}
\textbf{Sketch of proof.} First, we differentiate $e^2$ w.r.t. $\mathbf{t}$ and equate the partial derivative to 0. This gives us $\hat{\mathbf{t}} = (Y - RX) W\mathbf{1}$. Next, we substitute $X = KX + X\sqrt{\tilde{\mathbf{w}}}\sqrt{\tilde{\mathbf{w}}}^T$ on Eq.~\ref{eq:mse} and do the same for $Y$. Then, we substitute $\mathbf{t} = \hat{\mathbf{t}}$ into the squares. This yields $e^2 = \text{Tr}\left((Y-RX)KWK^T(Y-RX)^T\right)$ and expanding the term results in two squares plus $- 2\text{Tr}(YKWK^TX^TR^T)$. We maximize the last negative term whose maximum is the sum of all singular values, which leads to $\hat{R}$. The full derivation is in the supplement. \hfill\ensuremath{\square}
\vspace{0.5em}

We can easily extend the above theorem to incorporate a scaling factor $c \in \mathbb{R}^+$, or anisotropic scaling for tasks such as scan-to-CAD registration, but in this paper we assume that partial scans of the same scene have the same scale.

The Weighted Procrustes method generates rotation $\hat{R}$ and translation $\hat{\mathbf{t}}$ as outputs that depend on the weight vector $\mathbf{w}$. 
In our current implementation, $\hat{R}$ and $\hat{\mathbf{t}}$ are directly sent to the robust registration module in Section~\ref{sec:robust_registration} as an initial pose. 
However, we briefly demonstrate that they can also be embedded in an end-to-end registration pipeline, since Weighted Procrustes is differentiable.
From a top-level loss function $L$ of $\hat{R}$ and $\hat{\mathbf{t}}$, we can pass the gradient through the closed-form solver, and update parameters in downstream modules:
\begin{align}
    \frac{\partial}{\partial \mathbf{w}} L(\hat{R}, \hat{\mathbf{t}}) = & \frac{\partial L(\hat{R}, \hat{\mathbf{t}})}{\partial \hat{R}} \frac{\partial \hat{R}}{\partial \hat{\mathbf{w}}} + \frac{\partial L(\hat{R}, \hat{\mathbf{t}})}{\partial \hat{\mathbf{t}}} \frac{\partial \hat{\mathbf{t}}(\hat{R}, \hat{\mathbf{w}})}{\partial \hat{\mathbf{w}}},
    \label{eq:through_weighted_procrustes}
\end{align}
where $L(\hat{R}, \hat{\mathbf{t}})$ can be defined as the combination of differentiable rotation error (RE) and translation error (TE) between predictions $\hat{R}, \hat{\mathbf{t}}$ and ground-truth $R^*, \mathbf{t}^*$:
\begin{align}
     L_\text{rot}(\hat{R}) = & \arccos{\frac{\text{Tr}(\hat{R}^TR^*) - 1}{2}},  \label{eq:rte} \\
     L_\text{trans}(\hat{\mathbf{t}}) = & ||\hat{\mathbf{t}} - \mathbf{t}^*||_2^2, \label{eq:rre}
\end{align}
or the Forbenius norm of relative transformation matrices defined in~\cite{aoki2019pointnetlk,wang2019deep}.
The final loss is the weighted sum of $L_\text{rot}$, $L_\text{trans}$, and $L_\text{bce}$.
\section{Robust Registration}
\label{sec:robust_registration}
In this section, we propose a fine-tuning module that minimizes a robust loss function of choice to improve the registration accuracy.
We use a gradient-based method to refine poses, where a continuous representation~\cite{zhou2019continuity} for rotations is adopted to remove discontinuities and construct a smooth optimization space.
This module initializes the pose from the prediction of the Weighted Procrustes method.
During iterative optimization, unlike Maken \etal~\cite{maken2019speeding}, who find the nearest neighbor per point at each gradient step, we rely on the correspondence likelihoods from the 6D ConvNet, which is estimated only once per initialization.

In addition, our framework naturally offers a failure detection mechanism. In practice, Weighted Procrustes may generate numerically unstable solutions when the number of valid correspondences is insufficient due to small overlaps or noisy correspondences between input scans. By computing the ratio of the sum of the filtered weights to the total number of correspondences, \ie $\sum_i \phi(w_i) / |\mathcal{M}|$, we can easily approximate the fraction of valid correspondences and predict whether an alignment may be unstable.
When this fraction is low, we resort to a more time-consuming but accurate registration algorithm such as RANSAC~\cite{schnabel2007efficient,aiger20084pts, rusu2009icra} or a branch-and-bound method~\cite{yang2015go} to find a numerically stable solution.
In other words, we can detect when our system might fail before it returns a result and fall back to a more accurate but time-consuming algorithm, unlike previous end-to-end methods that use globally pooled latent features~\cite{aoki2019pointnetlk} or a singly stochastic matrix~\cite{wang2019deep} -- such latent representations are more difficult to interpret.

\newcommand{\ba}{\mathbf{a}}
\newcommand{\bb}{\mathbf{b}}
\newcommand{\bt}{\mathbf{t}}
\newcommand{\bx}{\mathbf{x}}
\newcommand{\by}{\mathbf{y}}
\subsection{SE(3) Representation and Initialization}
\label{sec:se3_representation}

We use the 6D representation of 3D rotation proposed by Zhou~\etal~\cite{zhou2019continuity}, rather than Euler angles or quaternions. The new representation uses 6 parameters $\ba_1, \ba_2 \in \mathbb{R}^3$ and can be transformed into a $3\times3$ orthogonal matrix by
\begin{equation}
f\left(
\begin{bmatrix}
| & | \\
\ba_1 & \ba_2 \\
| & |
\end{bmatrix}\right) = 
\begin{bmatrix}
| & | & | \\
\bb_1 & \bb_2 & \bb_3 \\
| & | & |
\end{bmatrix},
\label{eq:6d-representation}
\end{equation}
where $\bb_1, \bb_2, \bb_3 \in \mathbb{R}^3$ are $\bb_1 = N(\ba_1)$, $\bb_2 = N(\ba_2 - (\bb_1 \cdot \ba_2) \bb_1)$, and $\bb_3 = \bb_1 \times \bb_2$, and $N(\cdot)$ denotes L2 normalization.
Thus, the final representation that we use is $\ba_1, \ba_2, \bt$ which are equivalent to $R, \bt$ using Eq.~\ref{eq:6d-representation}.

To initialize $\ba_1, \ba_2$, we simply use the first two columns of the rotation matrix $R$, i.e., $\bb_1$, $\bb_2$. For convenience, we define $f^{-1}$ as $f^{-1}(f(R)) = R$ though this inverse function is not unique as there are infinitely many choices of $\ba_1, \ba_2$ that map to the same $R$.

\subsection{Energy Minimization}

We use a robust loss function to fine-tune the registration between predicted inlier correspondences. The general form of the energy function is
\begin{equation}
    E(R, \bt) = \sum_{i=1}^n \phi(w_{(i,J_i)}) L(\mathbf{y}_{J_i}, R\mathbf{x}_i + \bt),
    \label{eq:energy}
\end{equation}
where $\tilde{w}_i$ and $J_i$ are defined as in Eq.~\ref{eq:mse_w} and $\phi(\cdot)$ is a prefiltering function. In the experiments, we use $\phi(w) = I[w > \tau]w$, which clips weights below $\tau$ elementwise as neural network outputs bounded logit scores. $L(\bx, \by)$ is a pointwise loss function between $\bx$ and $\by$; we use the Huber loss in our implementation. 
The energy function is parameterized by $R$ and $\bt$ which in turn are represented as $\ba_1, \ba_2, \bt$. We can apply first-order optimization algorithms such as SGD, Adam, etc.\ to minimize the energy function, but higher-order optimizers are also applicable since the number of parameters is small.
The complete algorithm is described in Alg.~\ref{alg:dgr}. 
\begin{algorithm}[t]
\DontPrintSemicolon
\KwInput{$X \in \mathbb{R}^{n \times 3}, Y \in \mathbb{R}^{m \times 3}$} 
\KwOutput{$R \in \text{SO}(3), \bt \in \mathbb{R}^{3 \times 1}$}
    $\mathcal{F}_x \leftarrow \text{Feature}(X)$     
    \tcp*{\S~\ref{sec:features}}
    $\mathcal{F}_y \leftarrow \text{Feature}(Y)$\\
    $J_{x \rightarrow y} \leftarrow \text{NearestNeighbor}(\mathcal{F}_x, \mathcal{F}_y)$     \tcp*{\S~\ref{sec:6dconvnet}}
    $\mathcal{M} \leftarrow \{ (i,J_{x \rightarrow y, i}) \mid i \in [1, ..., n] \}$ \\
    $\mathbf{w} \leftarrow \text{InlierProbability}(\mathcal{M})$ \\
    
    \uIf{$\mathbb{E}_i \phi( w_i ) < \tau_s$}{ 
        \Return SafeGuardRegistration$(X, Y)$ \tcp*{\S\ref{sec:robust_registration}}
    }
    \Else{
            $\hat{R}, \hat{\bt} \leftarrow \argmin_{R, \bt} e^2(R, \bt; \mathbf{w}, X, Y)$    
        \tcp*{\S~\ref{sec:weighted_procrustes}}
        $\ba \leftarrow f^{-1}({\hat{R}}), \bt \leftarrow \hat{\bt}$
        \tcp*{\S~\ref{sec:se3_representation}}
        \While{\text{not converging}}
        {
            $\ell \leftarrow \sum_{(i,j) \in \mathcal{M}} \phi(w_{(i, j)}) L(Y_{j}, f(\ba) X_{i} + \bt)$\\
            $\ba \leftarrow \text{Update}(\ba, \frac{\partial}{\partial \ba} \ell(\ba, \bt))$\\
            $\bt \leftarrow \text{Update}(\bt, \frac{\partial}{\partial \bt} \ell(\ba, \bt))$
        }
        \Return $f(\ba), \bt$
    }
\caption{Deep Global Registration}\label{alg:dgr}
\end{algorithm}

\section{Experiments}
\label{sec:experiment}

We analyze the proposed model in two registration scenarios: pairwise registration where we estimate an $\text{SE}(3)$ transformation between two 3D scans or fragments, and multi-way registration which generates a final reconstruction and camera poses for all fragments that are globally consistent. Here, pairwise registration serves as a critical module in multi-way registration.

For pairwise registration, we use the 3DMatch benchmark~\cite{zeng20163dmatch} which consists of 3D point cloud pairs from various real-world scenes with ground truth transformations estimated from RGB-D reconstruction pipelines~\cite{HalberF2C17,dai2017bundlefusion}.
We follow the train/test split and the standard procedure to generate pairs with at least 30\% overlap for training and testing~\cite{ppf,ppf_fold,FCGF2019}.
For multi-way registration, we use the simulated Augmented ICL-NUIM dataset~\cite{choi2015cvpr,ahanda2014icra} for quantitative trajectory results, and Indoor LiDAR RGB-D dataset~\cite{Park2017} and Stanford RGB-D dataset~\cite{choi2015cvpr} for qualitative registration visualizations. Note in this experiment we use networks trained on the 3DMatch training set and do not fine-tune on the other datasets. This illustrates the generalization abilities of our models.
Lastly, we use KITTI LIDAR scans~\cite{kitti} for outdoor pairwise registration. As the official registration splits do not have labels for pairwise registration, we follow Choy~\etal~\cite{FCGF2019} to create pairwise registration train/val/test splits.

For all indoor experiments, we use 5cm voxel downsampling~\cite{rusu2009icra,Zhou2018}, which randomly subsamples a single point within each 5cm voxel to generate point clouds with uniform density. For safeguard registration, we use RANSAC and the safeguard threshold $\tau_s = 0.05$, which translates to 5\% of the correspondences should be valid.
We train learning-based state-of-the-art models and our network on the training split of the 3DMatch benchmark. During training, we augment data by applying random rotations varying from $-$180 to 180 degrees around a random axis. Ground-truth pointwise correspondences are found using nearest neighbor search in 3D space.
We train the 6-dimensional ConvNet on a single Titan XP with batch size 4. SGD is used with an initial learning rate $10^{-1}$ and an exponential learning rate decay factor 0.99.

\subsection{Pairwise Registration}%

\begin{figure*}[ht!]
    \centering
    \small    
    \includegraphics[width=.85\textwidth]{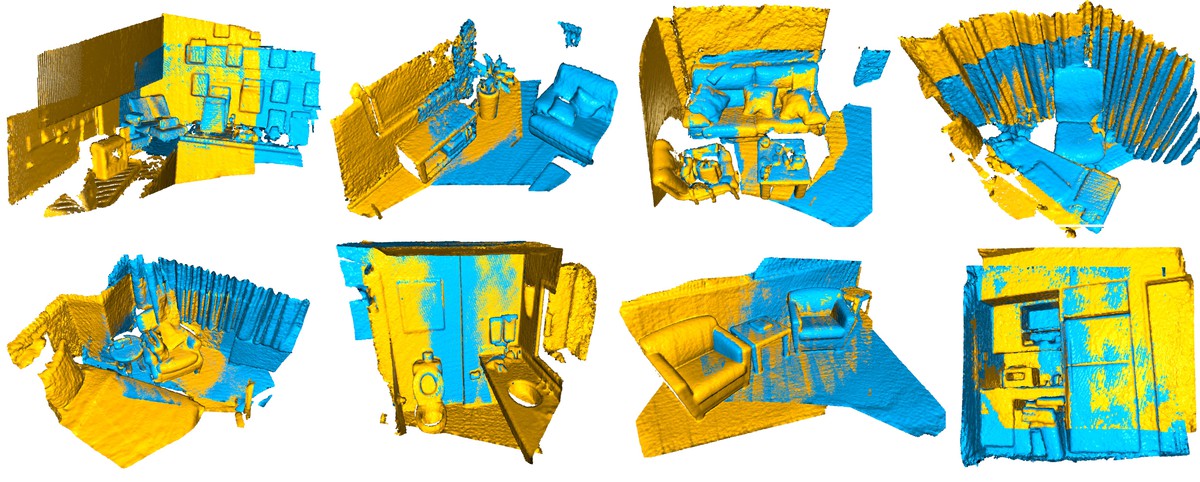}        
    \caption{Global registration results of our method on all 8 different test scenes in 3DMatch~\cite{zeng20163dmatch}. Best viewed in color.}
    \label{fig:8scenes}
\end{figure*}

\begin{figure*}
    \centering
    \includegraphics[width=.85\textwidth]{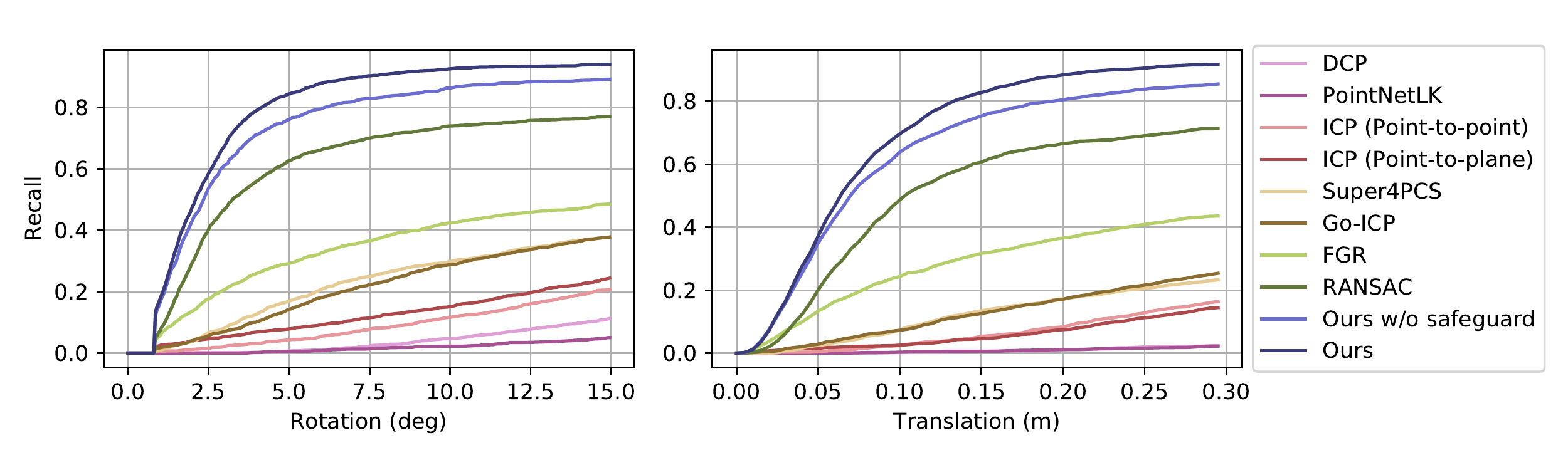}
    \small
    \caption{Overall pairwise registration recall (y-axis) on the 3DMatch benchmark with varying rotation (left image) and translation (right image) error thresholds (x-axis). Our approach outperforms baseline methods for all thresholds while being $6.5\times$ faster than the most accurate baseline.}
    \label{fig:3dmatch-recall}
\end{figure*}

In this section, we report the registration results on the test set of the 3DMatch benchmark~\cite{zeng20163dmatch}, which contains 8 different scenes as depicted in Fig.~\ref{fig:8scenes}.
We measure translation error (\textit{TE}) defined in Eq.~\ref{eq:rte}, rotation error (\textit{RE}) defined in Eq.~\ref{eq:rre}, and \textit{recall}. Recall is the ratio of successful pairwise registrations and we define a registration to be successful if its rotation error and translation error are smaller than predefined thresholds.
Average TE and RE are computed only on these successfully registered pairs since failed registrations return poses that can be drastically different from the ground truth, making the error metrics unreliable.

\begin{figure*}[ht!]
    \centering
    \small
    \includegraphics[width=.95\textwidth]{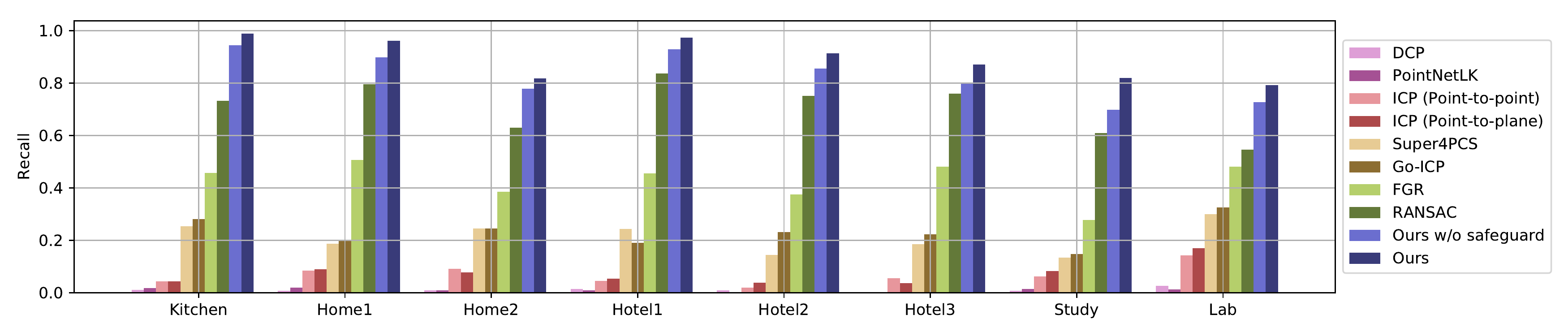}    
    \includegraphics[width=.95\textwidth]{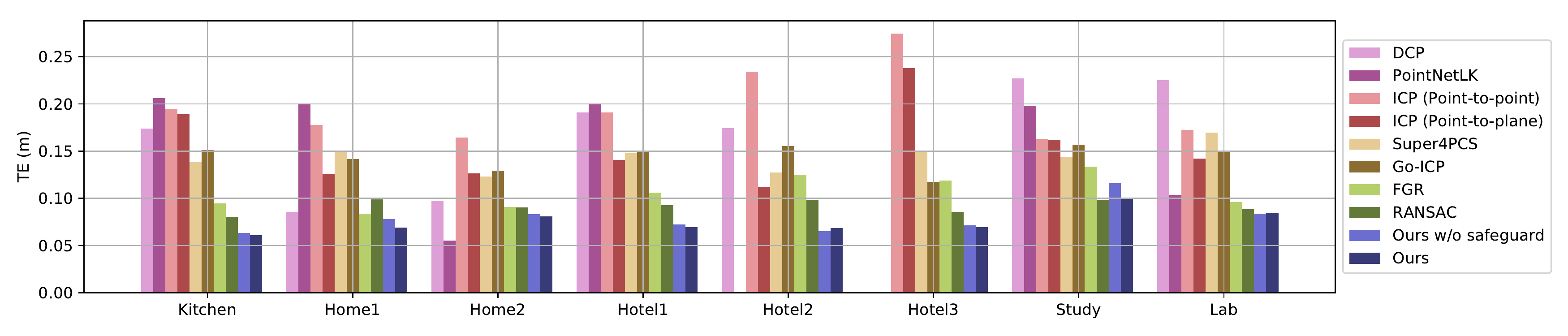}        
    \includegraphics[width=.95\textwidth]{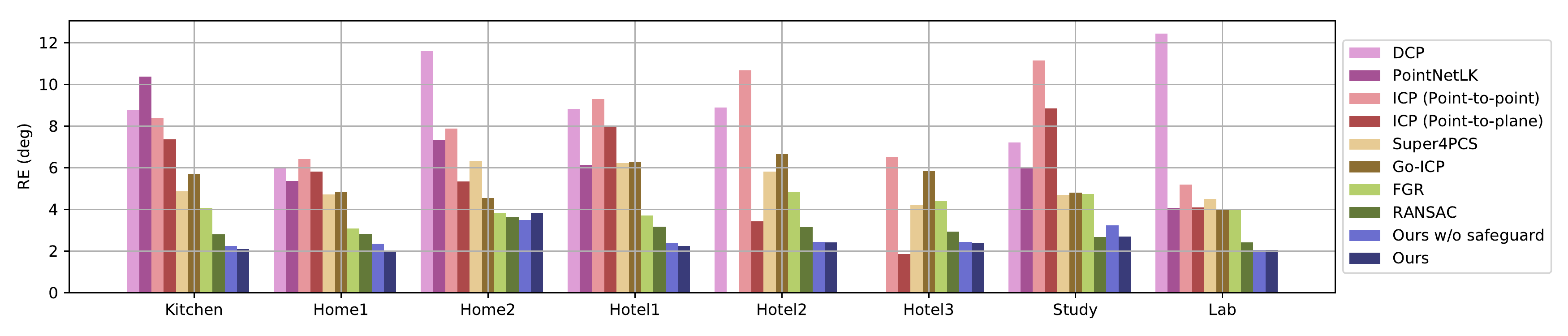}    
    \caption{Analysis of 3DMatch registration results per scene. \textit{Row 1}: recall rate (higher is better). \textit{Row 2-3}: TE and RE measured on successfully registered pairs (lower is better). Our method is consistently better on all scenes, which were not seen during training. Note: a missing bar corresponds to zero successful alignments in a scene.}
    \label{fig:scene-stats}
\end{figure*}

\begin{table}
\centering
\small
\caption{\textit{Row 1-6}: registration results of our method and classical global registration methods on point clouds voxelized with 5cm voxel size. Our method outperforms RANSAC and FGR while being as fast as FGR. \textit{Row 7-10}: results of ICP variants. \textit{Row 11 - 12}: results of learning-based methods. The learning-based methods generally fail on real-world scans.
Time includes feature extraction.}
\label{tab:abalation_voxel}
\resizebox{0.995\linewidth}{!}{
\begin{tabular}{l||c|c|c|c}
    \hline
     & Recall & TE (cm) & RE (deg) & Time (s) \\
    \hline
    Ours w/o safeguard            & {85.2}\% & {7.73} & {2.58} & 0.70 \\
    Ours                          & \textbf{91.3}\% & \textbf{7.34} & \textbf{2.43} & 1.21\\
    \hline
    FGR~\cite{zhou2016eccv}       & 42.7\% & 10.6 & 4.08 & \textbf{0.31} \\
    RANSAC-2M~\cite{rusu2009icra} & 66.1\% & 8.85 & 3.00 & 1.39 \\
    RANSAC-4M                     & 70.7\% & 9.16 & 2.95 & 2.32 \\
    RANSAC-8M                     & 74.9\% & 8.96 & 2.92 & 4.55 \\    
    \hline
    Go-ICP~\cite{yang2015go}             & 22.9\% & 14.7 & 5.38 & 771.0 \\
    Super4PCS~\cite{mellado2014super}    & 21.6\% & 14.1 & 5.25 & 4.55\\
    ICP (P2Point)~\cite{Zhou2018}        & 6.04\% & 18.1 & 8.25 & 0.25 \\    
    ICP (P2Plane)~\cite{Zhou2018}        & 6.59\% & 15.2 & 6.61 & 0.27 \\    
    \hline
    DCP~\cite{wang2019deep}              & 3.22\% & 21.4 & 8.42 & 0.07 \\    
    PointNetLK~\cite{aoki2019pointnetlk} & 1.61\% & 21.3 & 8.04 & 0.12 \\    
    \hline
\end{tabular}
}
\end{table}
\begin{table*}[h!]
\centering
\small
\caption{ATE (cm) error on the Augmented ICL-NUIM dataset with simulated depth noise. For InfiniTAM, the loop closure module is disabled since it fails in all scenes. For BAD-SLAM, the loop closure module only succeeds in `Living room 2'.}\label{table:ate-simulated}
\resizebox{0.995\linewidth}{!}{
    \begin{tabular}{l||c|c|c||c|c|c}
    \hline
    & ElasticFusion~\cite{Whelan2015} & InfiniTAM~\cite{Kahler2016} & BAD-SLAM~\cite{Schops2019} & Multi-way + FGR ~\cite{zhou2016eccv} & Multi-way + RANSAC ~\cite{Zhou2018}  & Multi-way + Ours \\
    \hline
    Living room 1 & 66.61 & 46.07 & fail & 78.97 & 110.9 & \textbf{21.06} \\%& 5.18 sec \\
    Living room 2 & 24.33 & 73.64  & 40.41 & 24.91 & \textbf{19.33} & 21.88 \\%& 3.19 sec \\
    Office 1      & \textbf{13.04} & 113.8 & 18.53 & 14.96 & 14.42 & 15.76 \\% & 4.96 sec \\
    Office 2     & 35.02 & 105.2 & 26.34 & 21.05 & 17.31 & \textbf{11.56} \\% & 3.72 sec \\
    \hline
    Avg. Rank & 3 & 5 & 5 & 3.5 & 2.5 & \textbf{2}\\
    \hline
    \end{tabular}
}

\end{table*}

\begin{figure}[h]
    \centering
    \subfloat[Real-world: \textit{Apartment}]{
    \includegraphics[width=.45\textwidth]{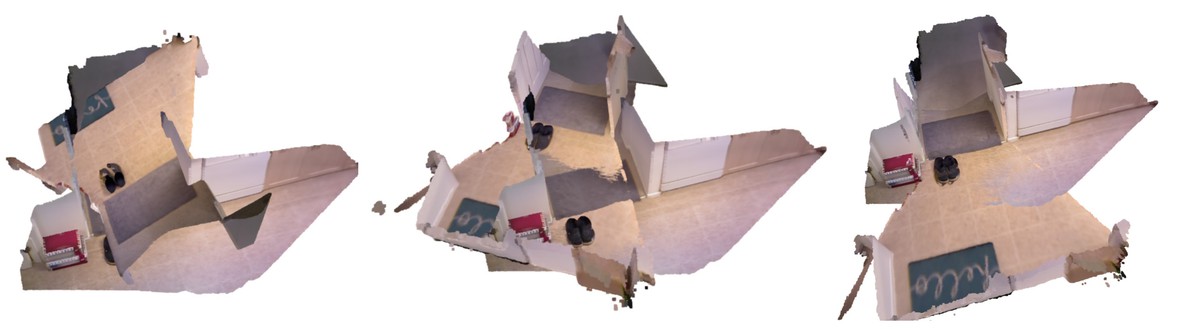}
    }\\
    \vspace{-0.7em}
    \subfloat[Real-world: \textit{Boardroom}]{
    \includegraphics[width=.45\textwidth]{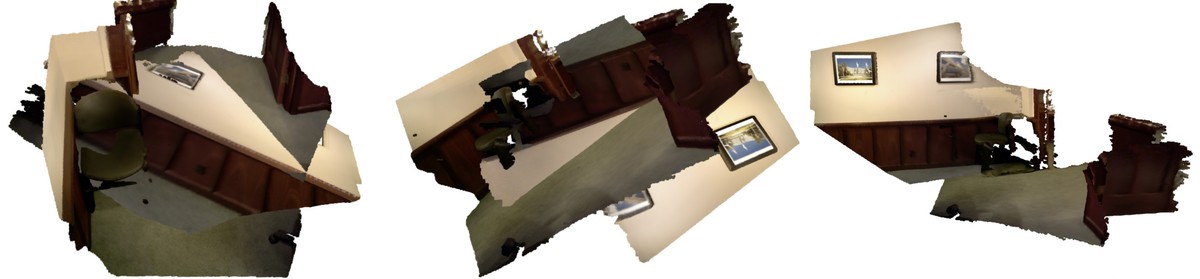}
    }\\ 
    \vspace{-0.9em}    
    \subfloat[Synthetic: \textit{Office}]{
    \includegraphics[width=.45\textwidth]{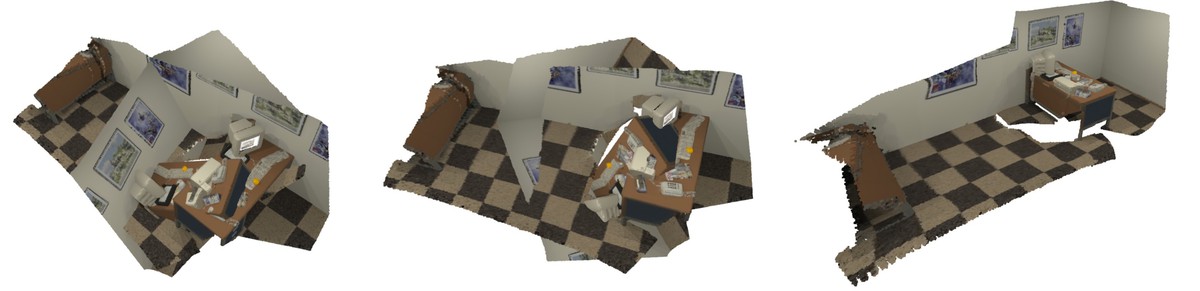}
    }
    \\
    \vspace{-0.7em}    
    \subfloat[Real-world: \textit{Copyroom}]{
    \includegraphics[width=.45\textwidth]{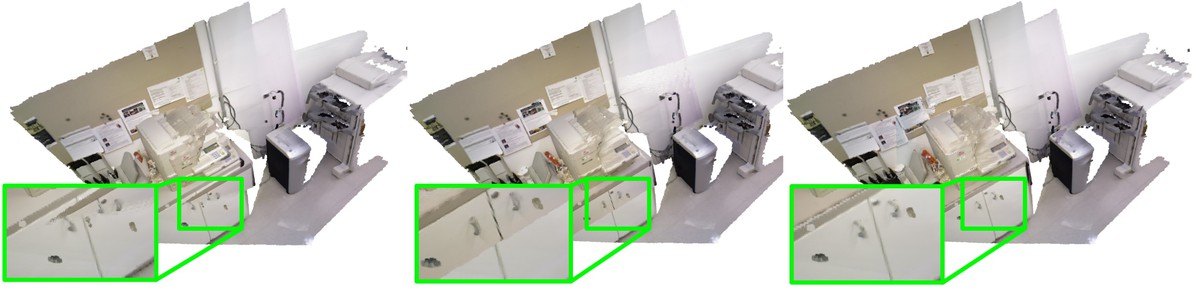}
    }\\
    \vspace{-0.7em}    
    \subfloat[Real-world: \textit{Loft}]{
    \includegraphics[width=.45\textwidth]{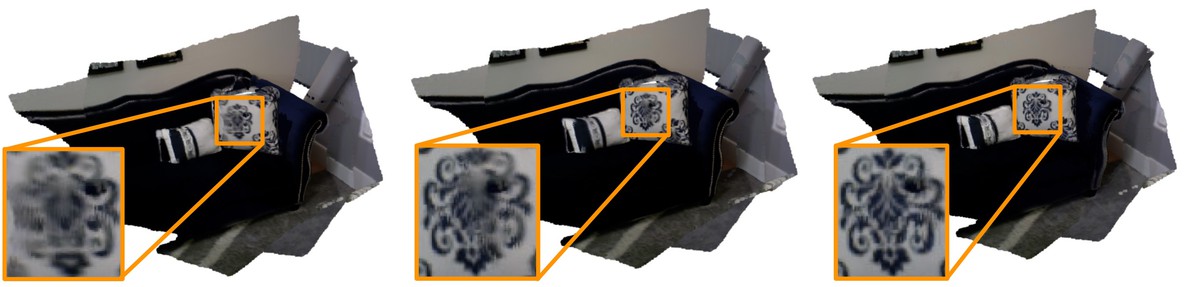}
    }\\ 
    \vspace{-0.7em}
    \subfloat[Synthetic: \textit{Livingroom}]{
    \includegraphics[width=.45\textwidth]{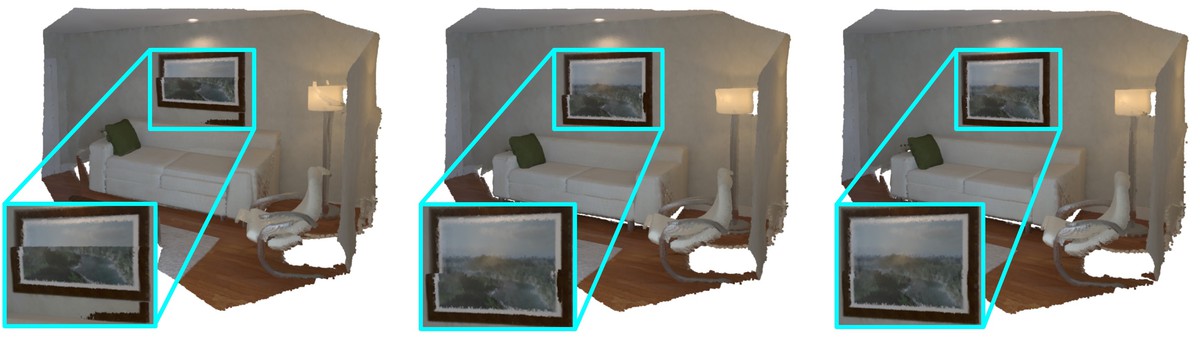}
    }
    \small
    \caption{Fragment registrations on~\cite{choi2015cvpr,Park2017}. From left to right: FGR~\cite{zhou2016eccv}, RANSAC~\cite{rusu2009icra}, Ours. \textit{Row 1-3}: our method succeeds on scenes with small overlaps or ambiguous geometry structures while other methods fail. \textit{Row 4-6}: by combining Weighted Procrustes and gradient-based refinement, our method outputs more accurate registrations in one pass, leading to better aligned details.}
    \label{fig:pairwise}
    \vspace*{-1em}
\end{figure}

We compare our methods with various classical methods~\cite{zhou2016eccv,rusu2009icra,yang2015go} and state-of-the-art learning based methods~\cite{wang2019deep,wang2019nips,aoki2019pointnetlk,dias2019corr}. 
All the experiments are evaluated on an Intel i7-7700
CPU and a GTX 1080Ti graphics card except for Go-ICP~\cite{yang2015go} tested on an Intel i7-5820K
CPU.
In Table~\ref{tab:abalation_voxel}, we measure recall with the TE threshold 30cm which is typical for indoor scene relocalization~\cite{murTRO2015}, and RE threshold 15 degrees which is practical for partially overlapping scans from our experiments. In Fig.~\ref{fig:3dmatch-recall}, we plot the sensitivity of recall on both thresholds by changing one threshold and setting the other to infinity. Fig.~\ref{fig:scene-stats} includes detailed statistics on separate test scenes. Our system outperforms all the baselines on recall by a large margin and achieves the lowest translation and rotation error consistently on most scenes.

\noindent \textbf{Classical methods.}
To compare with classical methods, we evaluate point-to-point ICP, Point-to-plane ICP, RANSAC~\cite{rusu2009icra}, and FGR~\cite{zhou2016eccv}, all implemented in Open3D~\cite{Zhou2018}. In addition, we test the open-source Python bindings of Go-ICP~\cite{yang2015go} and Super4PCS~\cite{mellado2014super}.
For RANSAC and FGR, we extract FPFH from voxel-downsampled point clouds.
The results are shown in Table~\ref{tab:abalation_voxel}.

ICP variants mostly fail as the dataset contains challenging 3D scan sequences with small overlap and large camera viewpoint change.
Super4PCS, a sampling-based algorithm, performs similarly to Go-ICP, an ICP variant with branch-and-bound search.

Feature-based methods, FGR and RANSAC, perform better. When aligning 5cm-voxel-downsampled point clouds, RANSAC achieves recall as high as 70\%, while FGR reaches 40\%.
Table~\ref{tab:abalation_voxel} also shows that increasing the number of RANSAC iterations by a factor of 2 only improves performance marginally.
Note that our method is about twice as fast as RANSAC with 2M iterations while achieving higher recall and registration accuracy.

\noindent \textbf{Learning-based methods.}
We use 3DRegNet~\cite{dias2019corr}, Deep Closest Point~(DCP)~\cite{wang2019deep}, PRNet~\cite{wang2019nips}, and PointNetLK~\cite{aoki2019pointnetlk} as our baselines. We train all the baselines on 3DMatch with the same setup and data augmentation as ours for all experiments.

For 3DRegNet, we follow the setup outlined in~\cite{dias2019corr}, except that we do not manually filter outliers with ground truth, and train and test with the standard realistic setup. %
We find that
the registration loss of 3DRegNet does not converge during training and the rotation and translation errors are consistently above 30 degrees and 1m during test.

We train Deep Closest Point (DCP) with 1024 randomly sampled points for each point cloud for 150 epochs~\cite{wang2019deep}. We initialize the network with the pretrained weights provided by the authors.
Although the training loss converges, DCP fails to achieve reasonable performance for point clouds with partial overlap.
DCP uses a singly stochastic matrix to find correspondences, but this formulation assumes that all points in point cloud $X$ have at least one corresponding point in the convex hull of point cloud $Y$. This assumption fails when some points in $X$ have no corresponding points in $Y$, as is the case for partially overlapping fragments.
We also tried to train PRNet~\cite{wang2019nips} on our setup, but failed to get reasonable results due to random crashes and high-variance training losses.

Lastly, we fine-tune PointNetLK~\cite{aoki2019pointnetlk} on 3DMatch for 400 epochs, starting from the pretrained weights provided by the authors.
PointNetLK uses a single feature that is globally pooled for each point cloud and regresses the relative pose between objects, and we suspect that a globally pooled feature fails to capture complex scenes such as 3DMatch.

In conclusion, while working well on object-centric synthetic datasets, current end-to-end registration approaches fail on real-world data. %
Unlike synthetic data, real 3D point cloud pairs contain multiple objects, partial scans, self-occlusion, substantial noise, and may have only a small degree of overlap between scans.

\subsection{Multi-way Registration}
Multi-way registration for RGB-D scans proceeds via multiple stages. First, the pipeline estimates the camera pose via off-the-shelf odometry and integrates multiple 3D scans to reduce noise and generate accurate 3D fragments of a scene. Next, a pairwise registration algorithm roughly aligns all fragments, followed by multi-way registration~\cite{choi2015cvpr} which optimizes fragment poses with robust pose graph optimization~\cite{kummerle2011g}.

We use a popular open-source implementation of this registration pipeline~\cite{Zhou2018} and replace the pairwise registration stage in the pipeline with our proposed modules.
Note that we use the networks trained on the 3DMatch training set and test on the multi-way registration datasets~\cite{ahanda2014icra,Park2017,choi2015cvpr}; this demonstrates cross-dataset generalization.

We test the modified pipeline on the Augmented ICL-NUIM dataset~\cite{choi2015cvpr,ahanda2014icra} for quantitative trajectory results, and Indoor LiDAR RGB-D dataset~\cite{Park2017} and Stanford RGB-D dataset~\cite{choi2015cvpr} for qualitative registration visualizations.
We measure the absolute trajectory error (ATE) on the Augmented ICL-NUIM dataset with simulated depth noise. As shown in Table~\ref{table:ate-simulated}, compared to state-of-the-art online SLAM~\cite{Whelan2015,Kahler2016,Schops2019} and offline reconstruction methods~\cite{zhou2016eccv}, our approach yields consistently low error across scenes.

For qualitative results, we compare pairwise fragment registration on these scenes against FGR and RANSAC in Fig.~\ref{fig:pairwise}. Full scene reconstruction results are shown in the supplement.

\subsection{Outdoor LIDAR Registration}

\begin{figure}[ht!]
    \centering
    \subfloat[KITTI registration test pair 1]{
    \includegraphics[width=.95\linewidth]{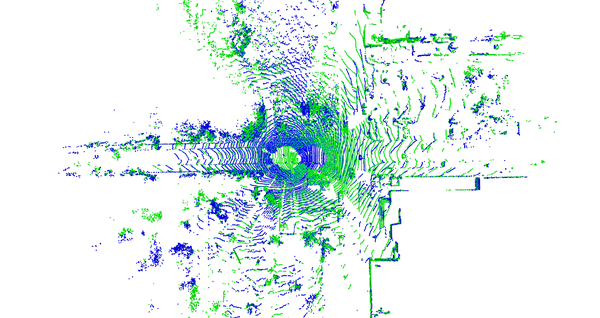}
    }\\
    \subfloat[KITTI registration test pair 2]{
    \includegraphics[width=.95\linewidth]{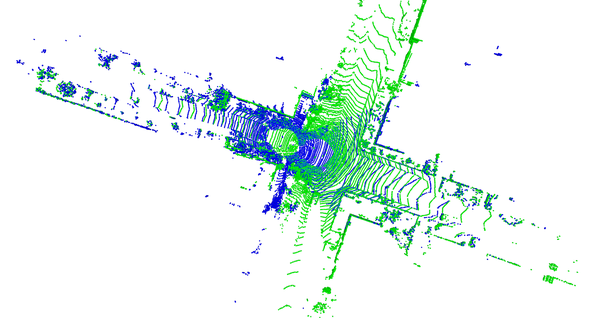}
    }\\
    \subfloat[KITTI registration test pair 3]{
    \includegraphics[width=.95\linewidth]{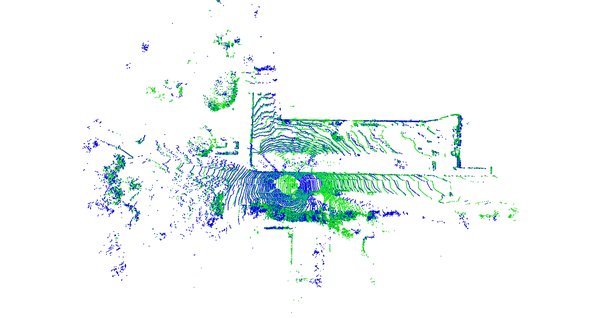}
    }\\
    \caption{KITTI registration results. Pairs of scans are at least 10m apart.}
    \label{fig:kitti-registration}
\end{figure}

We use outdoor LIDAR scans from the KITTI dataset~\cite{kitti} for registration, following~\cite{FCGF2019}. The registration split of Choy~\etal~\cite{FCGF2019} uses GPS-IMU to create pairs that are at least 10m apart and generated ground-truth transformation using GPS followed by ICP to fix errors in GPU readings.
We use FCGF features~\cite{FCGF2019} trained on the training set of the registration split to find the correspondences and trained the 6D ConvNet for inlier confidence prediction similar to how we trained the system for indoor registration.
We use voxel size 30cm for downsampling point clouds for all experiments. Registration results are reported in Tab.~\ref{tab:kitti_comparison} and visualized in Fig.~\ref{fig:kitti-registration}. 
\begin{table}
\centering
\small
\caption{Registration on the KITTI test split~\cite{kitti,FCGF2019}. We use thresholds of 0.6m and 5 degrees. `Ours + ICP' refers to our method followed by ICP for fine-grained pose adjustment. The runtime includes feature extraction.}
\label{tab:kitti_comparison}
\resizebox{0.995\linewidth}{!}{
\begin{tabular}{l||c|c|c|c}
    \hline
                                    & Recall & TE (cm) & RE (deg) & Time (s) \\
    \hline
    FGR~\cite{zhou2016eccv}         & 0.2\%  & 40.7 & 1.02 & 1.42 \\
    RANSAC~\cite{rusu2009icra}      & 34.2\% & 25.9 & 1.39 & 1.37 \\
    FCGF~\cite{FCGF2019}            & \textbf{98.2\%} & 10.2 & 0.33 & 6.38 \\
    \hline
    Ours                            & 96.9\% & 21.7 & 0.34 & \textbf{2.29} \\
    Ours + ICP                      & 98.0\% & \textbf{3.46} & \textbf{0.14} & 2.51 \\
    \hline
\end{tabular}
}
\end{table}

\section{Conclusion}

We presented Deep Global Registration, a learning-based framework that robustly and accurately aligns real-world 3D scans. To achieve this, we used a 6D convolutional network for inlier detection, a differentiable Weighted Procrustes algorithm for scalable registration, and a gradient-based optimizer for pose refinement. Experiments show that our approach outperforms both classical and learning-based registration methods, and can serve as a ready-to-use plugin to replace alternative registration methods in off-the-shelf scene reconstruction pipelines.

{\small
\bibliographystyle{ieee_fullname}
\bibliography{egbib}
}

\end{document}